\documentclass[a4paper,twocolumn,10pt]{article}%{grantapp}

\usepackage{cite}
\usepackage{graphicx}

\usepackage{xcolor}

\usepackage{amsmath,amssymb}
\usepackage{url}
\newcommand{\etal}{\textit{et al}.}
\usepackage{balance}

\begin{document}
%\title{Full Mesh 3D Human Pose Recovery from Monocular Video}

\title{Temporally Coherent Full 3D Mesh Human Pose Recovery from Monocular Video} 
\author{Jian Liu, Naveed Akhtar and Ajmal Mian}
\date{}

\maketitle

\begin{abstract}

Advances in Deep Learning have recently made it possible to recover full 3D meshes of human poses from individual images. However, extension of this notion to videos for recovering  temporally coherent poses still remains unexplored. A major challenge in this regard is the lack of appropriately annotated video data for learning  the desired deep models. Existing human pose datasets only provide 2D or 3D skeleton joint annotations, whereas the datasets are also recorded in constrained environments.
We first contribute a technique to synthesize monocular action videos with rich 3D annotations that are suitable for learning computational models for full mesh 3D human pose recovery. 
Compared to the existing methods which simply ``texture-map'' clothes onto the 3D human pose models, 
%Compared to the existing ``texture-mapped'' human pose synthesis methods, 
our approach incorporates Physics based realistic cloth deformations with the human body movements. 
The generated videos cover a large variety of human actions, poses, and visual appearances, whereas the  annotations record accurate human pose dynamics and human body surface information.
Our second major contribution is an end-to-end trainable Recurrent Neural Network for full pose mesh recovery from monocular video.
Using the proposed video data and LSTM  based recurrent structure, our network explicitly learns to model the temporal coherence in videos and imposes geometric consistency over the recovered meshes. We establish the effectiveness of the proposed model with quantitative and qualitative analysis using the proposed and benchmark datasets.

%Deep learning has recently demonstrated the possibility of recovering full 3D meshes for human poses from individual images. 
%However, extension of this concept to videos to recover temporally coherent poses still remains largely unexplored. One major challenge in this regard is the lack of appropriately annotated video data. Existing human pose datasets include only 2D or 3D joints annotations, and most of them are recorded in constrained environments.
%We first contribute a method to synthesize monocular action videos with rich 3D annotations that are suitable for pose mesh recovery learning. Compared to previous ``texture-mapping'' human pose synthesis methods, our technique incorporates Physics based realistic cloth deformations with body movement. 
%The generated videos cover a large variety of human actions, poses, and visual appearances, and the annotations record accurate human pose dynamics and human body surface information. 
%Secondly, we contribute an end-to-end trainable Recurrent Neural Network for full pose mesh recovery from monocular video. 
%With the proposed video data and the recurrent structure, our network explicitly models the temporal coherence in videos, and imposes geometric consistency over the recovered meshes. Quantitative and qualitative experiments with the proposed and benchmark datasets establish the effectiveness of our technique.

\end{abstract}

%\begin{IEEEkeywords}
\noindent{\bf Keywords:} Human Pose Recovery, 3D Human Reconstruction, Full Mesh Recovery, Data Synthesis.
%\end{IEEEkeywords}

% \fi
%\IEEEpeerreviewmaketitle

\section{Introduction}
\label{sec:Intro}
Recovering human poses from monocular (as~opposed~to~multiview) images is an efficient approach since it does not require cumbersome calibration or high cost equipment. It has many applications in pose transfer, human movement analysis and action recognition. Until recently, the techniques used for human pose recovery aimed at predicting skeletal joint configurations from  images~\cite{wei2016convolutional,mehta2017vnect,pavlakos2017coarse,wang2018drpose3d,luo2018lstm}. However, recent findings~\cite{kanazawa2018end,yang20183d} ascertain that,  using deep learning, it is possible to reconstruct  full 3D human meshes from monocular images with the help of  parameterized body and shape configurations~\cite{loper2015smpl}. 
Full 3D mesh recovery has clear advantages over the sparse skeleton recovery of human poses, as the former captures the inner pose dynamics as well as the outer 3D human bodies. These advantages multiply  when, instead of individual images, human meshes can be recovered for full videos while incorporating the temporal dynamics of the body movement.

\begin{figure}[t]
\centering
\includegraphics[width=0.48\textwidth]{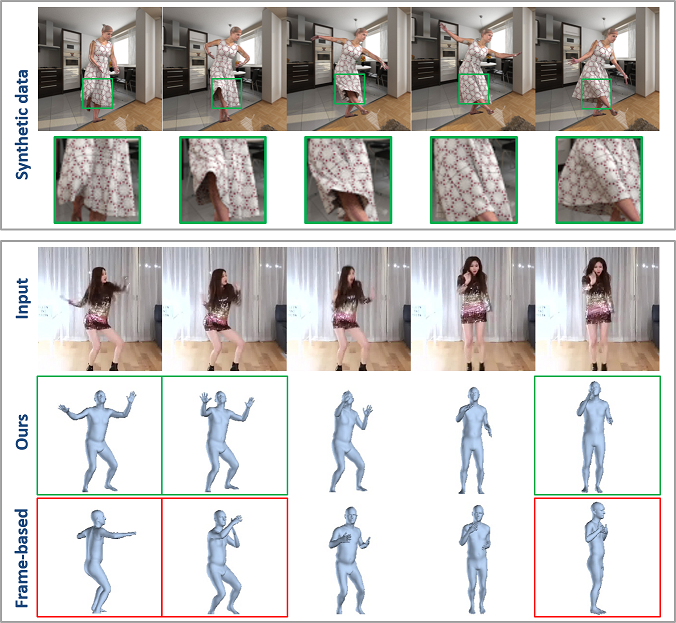}
\caption{(Top) Illustration of realism in our data due to Physics based cloth movement. (Bottom) Compared to conventional frame-based 3D pose recovery~\protect\cite{kanazawa2018end}, the proposed method better models the temporal variation and enforces geometric consistency of poses. The highlighted frames show frames where the difference between the two techniques is significant.}
\label{fig:teaser}
%\vspace{-3mm}
\end{figure}

Video based full mesh recovery of human poses has further applications in precision modeling of human actions, virtual try-on, automatic animation, human-computer interaction and so on. However, progress in this research direction is currently hampered by the unavailability of appropriately annotated data for learning the desired computational models. Due to the constraints of sensor specifications, data modality and data sample size; existing datasets for human pose recovery, e.g.~\cite{lin2014microsoft,h36m_pami,mehta2017vnect} are not particularly helpful in recovering  full mesh poses from videos. On the other side, the requirement of panoptic studios~\cite{joo2015panoptic} or full body scanners{~\cite{pons2017clothcap,mahmood2019amass}} to annotate data for this task restrict  researchers in generating specialized data for their specific problems in this area.

This work first addresses the problem of generating appropriately annotated training data to learn  computational models that can recover full 3D meshes of human poses from monocular videos. 
To that end, we introduce a video generation technique that provides rich 3D annotations for realistic monocular videos. The generated videos are able to easily incorporate a wide variety of human actions and other visual variations related to e.g.~cloth-body and cloth-gravity interactions, cloth textures, lighting conditions, camera viewpoints, and scene backgrounds.
The annotations recorded for these videos include parameters of 3D human avatar (shape), 3D skeleton (pose) and its 2D image projection, and even the vertices of full human meshes (pose) in the videos.
To accurately capture the movements of clothes and their interaction with human bodies, we employ a Physics engine that endows the generated videos with fine-grained realistic effects. 
%{\color{blue}The recorded annotations include parameters of 3D human avatar, 3D skeleton and its 2D image projection, and even vertices of the full human mesh.
%To simulate physical effects of cloth and its reaction with human body in the generated videos,} we employ a Physics engine to accurately capture the cloth movement with actions.
To the best of our knowledge, this is the first of its kind ability of a data generation technique in the broader research direction of video/image based human action analysis. The videos/frames  resulting from our technique are much more realistic as compared to those generated by the approaches that simulate clothes as textures on human models, see Fig.~\ref{fig:teaser}-top. The source code of our technique and the resulting data will be made public for the broader research community.

The second major contribution of this work is an end-to-end trainable Recurrent Neural Network that recovers  full 3D human pose meshes from monocular video. The proposed network embeds a foundational building block that processes an individual video frame in a recurrent structure. We incorporate attention mechanism in the recurrent structure that  provides contextual conditioning based on low-level visual features. The network precisely models the spatio-temporal dynamics of human movements in videos and imposes geometric consistency over the recovered mesh sequences by incorporating a body shape smoothness loss in the training process.

We analyze our technique using the proposed data and the benchmark Human3.6M~\cite{h36m_pami} and UCF101~\cite{soomro2012ucf101} datasets.
Our experiments demonstrate that the  proposed method is able to achieve very promising full mesh recovery for pose estimation from monocular video, e.g.~Fig.~\ref{fig:teaser}-bottom. Due to the nascency of this research direction, this article also makes a minor contribution in introducing three new evaluation metrics to more appropriately analyze the mesh pose recovery from videos as compared to joint based pose recovery metrics used for individual video frames.

\section{Related Work}
\label{sec:RW}

Traditional techniques of  human pose recovery usually apply  pictorial structure models~\cite{andriluka2009pictorial,felzenszwalb2005pictorial,sapp2011parsing,yang2013articulated,andriluka2010monocular} to optimize body part configurations. However, due to the rapid advancements in deep learning, recovering articulated human poses using Convolutional Neural Network (CNN) models has become increasingly popular. For instance, Wei \etal~\cite{wei2016convolutional} proposed Convolutional Pose Machines (CPM) to estimate 2D pose keypoints by learning a multi-stage CNN. The authors used a sequential convolutional structure to capitalize on the spatial context and iteratively updated the belief maps. In their method, the receptive field of neurons is carefully designed at each stage to allow the learning of complex and long-range correlations between the body parts. Similarly, Newell \etal~\cite{newell2016stacked} proposed a  Stacked Hourglass method to process visual features across different scales. Their results are consolidated to better capture various spatial relationships  associated with human body. Nevertheless, the CPM and Stacked Hourglass only work for static frame/images and do not account for  any geometric consistency between different video frames.

To estimate 2D human poses in video, Pfister~\etal ~\cite{pfister2015flowing} exploited the  temporal context in video by combining information across multiple frames with optical flow. They used the resulting information to align heatmap predictions from the neighbouring frames. As a more recent attempt in modeling temporal information for human pose recovery,
Luo~\etal~\cite{luo2018lstm} re-modelled CPM as a Recurrent Neural Network to replace the multiple stages of CPM with sequential LSTM cells. The concept of using hand-crafted optical flow or recurrent structure to model temporal information in pose recovery task is beneficial. Nevertheless, both of these methods remain limited to recovering 2D keypoints only.

It is challenging to extend 2D keypoints recovery methods to recover 3D skeletons, as the latter demands sophisticated solutions. For example, Camillo~\cite{taylor2000reconstruction} had to enforce additional constraints on the relative lengths of human limbs and the body joint kinematics to select valid limb configuration. Ramakrishna \etal~\cite{ramakrishna2012reconstructing} proposed an activity-independent technique to recover 3D joint  configurations  using 2D locations of anatomical landmarks. They also leveraged a large motion capture corpus as a proxy to infer the plausible 3D configurations. A few contributions in this direction have also formulated 3D skeleton recovery as a supervised learning problem. For instance, Pavlakos~\etal~\cite{pavlakos2017coarse} proposed a volumetric technique to estimate 3D human poses from a single image. Their volumetric representation converts the 3D coordinate regression problem to a more manageable prediction task in a discretized space. With this representation, they concatenated multiple fully convolutional network components to implement a iterative coarse-to-fine learning process.

Mehta~\etal~\cite{mehta2017monocular} enhanced CNN supervision with intermediate heat-maps, and used transfer learning from in-the-wild 2D pose data to improve the generalization to in-the-wild images for 3D pose recovery task. As a typical per-frame pose estimation technique, their method exhibits temporal jitters in video
sequences. Another technique, VNect~\cite{mehta2017vnect} formulates the 3D skeleton recovery problem as a CNN pose regression task that follows an optimization process,  termed kinematic skeleton fitting.  It is specifically designed to  improve temporal stability of the recovered poses.
Wang~\etal~\cite{wang2018drpose3d} proposed a two-step technique for 3D pose estimation, named DRPose3D. In the first step, it uses a Pairwise Ranking CNN to extract depth rankings of human joints from images. In the second step, it uses this information to regress the 3D poses. This method  depends on a 2D pose estimator for computing the initial joint heat maps. Consequently, it can not be treated as an end-to-end trainable technique.

Due to its attractive applications, full 3D mesh pose recovery from images is an emerging direction. {Many recent methods adopt the parametric human model~\cite{loper2015smpl} in this regard. For instance, Alldieck~\etal~\cite{alldieck2018video,alldieck2018detailed,alldieck2019learning} inferred 3D shape of person with details including hair and cloth using parametric model.  Yu~\etal~\cite{yu2018doublefusion,yu2019simulcap} also used depth sensor to reconstruct 3D body shapes in cloth. They used cloth simulation for single view human performance capture.}
Kanazawa~\etal~\cite{kanazawa2018end} adopted the parametric human model for 3D mesh pose recovery and predicted its pose and shape parameters from monocular images in an end-to-end manner.
Such a learning-based method requires training data with 3D annotations~\cite{omran2018neural}, a requirement fulfilled by very few datasets~\cite{h36m_pami,sigal2010humaneva}. 
The lack of training data has also driven research to exploit generative adversarial networks for 3D pose learning. Kanazawa~\etal~\cite{kanazawa2018end} used unpaired 3D annotations to create a factorized adversarial prior. Similarly, Yang~\etal~\cite{yang20183d} learned a discriminator to enforce the 3D pose estimator to generate the plausible poses. %Adversarial priors learned from un-annotated data suffer from bias in the data distribution.
However, a major limitation of such methods is that their complexity increases significantly when moving  from frame to video modeling.

%These adversarial-related methods mitigates the lack of 3D training data, however, the adversarial prior learned from the chosen un-annotated dataset suffers from bias in data distribution. Another limitation with adversarial prior is the increased model complexity when shifting from frame-based model to video-based framework.

The challenges in manual annotation of large-scale data has also led researchers to synthesize annotated data. For instance, {Lassner~\etal~\cite{lassner2017generative} proposed a generative model to create people and manipulate  their clothes. However, their method is frame-based which compromises the dynamic details of the clothes.}
Varol~\etal~\cite{varol17} proposed SURREAL to synthesize human images for the tasks of body segmentation and depth estimation. %The generated dataset includes RGB images, depth maps and body parts segmentation information. The 3D joints annotations are also available in this method. 
%Firstly, the variations of human actions are limited in SURREAL.
Although SURREAL is able to generate RGB human images and 3D joints annotations it has multiple  shortcomings. For instance, the variations of human actions are limited in SURREAL. Moreover, the images in the  dataset remain unrealistic in terms of interaction between human models and their clothes. This happens because the data generation method  simply wraps 2D cloth textures onto human models.
Our literature survey reveals that  end-to-end learning for 3D pose mesh recovery is a promising direction but remains largely unexplored due to the unavailability of large-scale data with rich 3D annotations. Moreover, due to single frame based benchmark datasets, existing methods are also limited to process individual frames. We address these issues by proposing a data generation method and a temporally coherent technique for full 3D mesh human pose recovery from videos.   %This work is aimed at removing this bottleneck, and also demonstrating the possibility of tapping into 

%\vspace{-2mm}
\section{Data Generation}
\label{sec:dataGen}
As the first major contribution of this work, below we introduce our method of computationally generating human action videos with rich 3D annotations for training. 
%The resulting data enables full mesh recovery of human poses from videos.
%Our method adopts the human model SMPL~\cite{loper2015smpl}} to create avatars, and uses CMU MoCap database~{\color{red}[ref]} to endow avatars with a variety of poses and motions. {{\color{red}a sentence more on what we do.}}
%In this section, we explain our workflow to synthesize human action video dataset. The workflow adopts the parametric human model SMPL to create avatars, and uses CMU MoCap database to endow avatars with a variety of poses and motions.

%\vspace{-2mm}
\subsection{Human Pose and Shape Model}
\label{sec:SMPL}
To represent human avatars, we use the Skinned Multi-Person Linear (SMPL) model \cite{loper2015smpl} as parametric representation of human avatars in our data generation pipeline. 
SMPL provides a skinned vertex-based representation that can encode a wide variety of human body shapes in natural poses.
The fact that SMPL model is created statistically using a large number of \textit{real humans} also makes it suitable for the pose recovery task. 
An avatar in SMPL representation is given as a tuple $\mathcal{M}(\boldsymbol\beta, \boldsymbol\theta)$, where $\boldsymbol\beta$ and $\boldsymbol\theta$ respectively encode the body shape and pose.  The body shape is parameterized by the first 10 coefficients of shape PCA space, hence $\boldsymbol\beta \in \mathbb{R}^{10}$. For $\boldsymbol{\theta}$, selected bones in human skeleton are represented in  a hierarchical tree, where each bone/node is connected to its parent, and the whole skeleton is anchored to a root node. At each body joint $j$, an axis-angle rotation vector $\boldsymbol{\rho}_j \in \mathbb{R}^{3}$ controls the rotation of a child bone relative to its parent bone. The orientation of whole body is controlled by the root rotation vector $ \boldsymbol{\rho}_0 \in \mathbb{R}^{3}$. All of the above bone kinematics information is summarized by the pose parameter $\boldsymbol\theta \in \mathbb{R}^{3K}$, where $K=23$ for the $23$ body joints chosen by the SMPL representation. %, plus the root node. %$\boldsymbol\gamma \in \mathbb{R}^{3}$ encodes the translation of the root node.

In terms of SMPL representation, an avatar's pose ${\bf P}_i \in \mathbb{R}^{3 \times N}$ is a mapping of a tuple $\mathcal{M}_i(\boldsymbol\beta_i, \boldsymbol\theta_i)$ to $N=6890$ vertices describing the surface of an avatar. It is also possible to extract 2D keypoints and 3D skeletons  using the technique of \cite{loper2015smpl}. Our method uses the SMPL representation to record an avatar's pose that is later rendered to a video frame using the graphics pipe-line discussed below.

%Along with the generation of synthetic videos, we record the parameter $\beta, \theta$ as the ground truth of SMPL models. In addition, it is convenient to get the 3D joints/skeleton and 3D vertex of SMPL models. In the 3D human pose recovery, we train a neural network to estimate SMPL parameters $\beta, \theta$ directly from RGB input videos. With the estimated parameters, SMPL model is able to output triangulated surface, which can be used to derive predicted 3D joints/skeleton. As SMPL model $M$ is differentiable, we then use the predicted 3D joints to calculate and back propagate the 3D joints fitting error for the neural network training. The detailed training process will be elaborated in later sections.

%\vspace{-1mm}
\subsection{Model Pose Variations}
We exploit the CMU MoCap database (\url{http://mocap.cs.cmu.edu}) to bequeath the SMPL avatars with a large variety of natural poses and motions recorded using real humans. The CMU MoCap dataset covers more than $2500$  different action sequences that capture the dynamics of 3D skeleton joints. We use the MoSh technique~\cite{loper2014mosh} to map CMU joint locations to SMPL model parameter - resulting in human  avatars. 
For a given CMU MoCap sequence, MoSh estimates the SMPL parameter $\boldsymbol\theta$ that best explains the body joint rotations corresponding to the CMU skeleton data.
Multiple SMPL $\boldsymbol\beta$ parameters can then be chosen that animate the same action under different body shapes. We call the thus generated  SMPL sequences  as ``MoShed'' sequences. % in the text to follow. 
%{\color{red} We empirically selected multiple $\boldsymbol\beta$ and $\boldsymbol\gamma$ to generate XXX MoShed sequences in this work.}

\begin{figure}[t]
\centering
\includegraphics[width=0.48\textwidth]{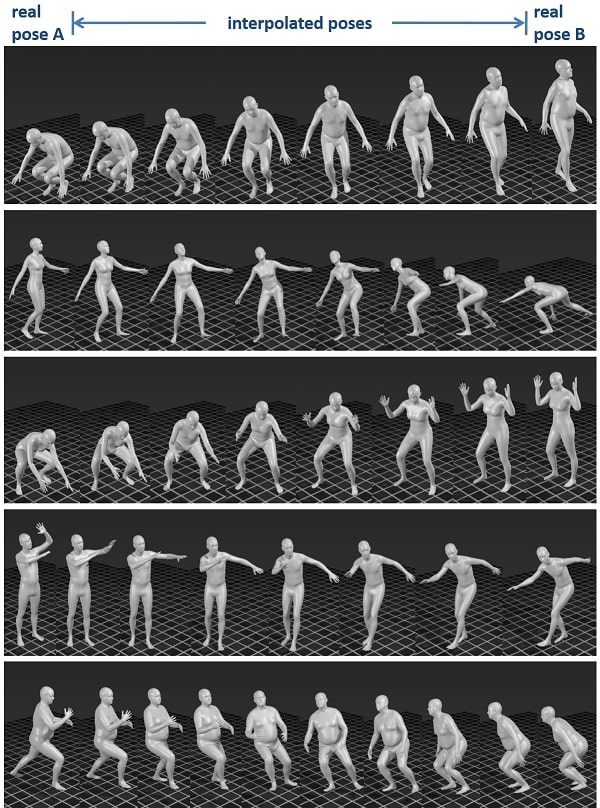}
\caption{Illustration of pose interpolation: The first and last pose in each row are original poses, denoted as A and B. All the remaining poses in a row are generated by interpolating between their real pose A and B.}
\label{fig:pose_interpolation}
%\vspace{-3mm}
\end{figure}

When using the MoShed CMU sequences, the pose variations are upper bounded by the total  CMU pose types. To enhance the pose variations, we employ a pose interpolation technique to create new MoShed sequences that consist of novel poses. While interpolating between two poses, we choose widely contrasting poses as the starting and ending pose. See the real poses A and B in Fig.~\ref{fig:pose_interpolation} as representative examples. This choice results in the interpolated poses that are significantly different from the available real pose sequences in the CMU dataset, improving the pose variety in our dataset.

%We propose a pose interpolation approach to expand existing CMU MoCap action sequences. 

The selection of contrasting real poses for the interpolation is made as follows. Consider two MoShed CMU MoCap sequences ${\bf X} = \{X_i\}_{i = 1}^m$  and ${\bf Y} = \{Y_j\}_{j = 1}^n$ that respectively contain $m$ and $n$ frames. Each frame in these sequences contains an independent human pose that is represented by SMPL pose parameter $\boldsymbol\theta$. 
We define the distance between a pair of human poses as $dist_{i,j}=||X_i (\boldsymbol\theta) - Y_j (\boldsymbol\theta) ||_2$, and compute a distance matrix ${\bf D} \in \mathbb{R}^{m\times n}$ for all the pose pairs for action sequences ${\bf X}$ and ${\bf Y}$ using this distance. We select the two poses for interpolation by getting the frame indices $i,j = \text{argmax}({\bf D})$, and record $dist_{i,j}=\text{max}({\bf D})$. %The two resulting human poses are chosen as the most distinctive pose pairs in the sequences ${\bf X}$ and ${\bf Y}$.
These pairs are used as the starting and ending frames for the creation of new pose sequence through interpolation.

As mentioned in Section~\ref{sec:SMPL}, $\boldsymbol\theta \in \mathbb{R}^{3K}$ represents axis-angle rotation for each body joint relative to its parent bone. Compared to Quaternion rotation, Axis-angle rotation normalizes the rotation axis and multiplies it with the rotation magnitude. As $\boldsymbol\theta$ represents relative rotation and it works independently on each body joint, it is convenient to perform linear pose interpolation using the $\boldsymbol\theta$ parameters of the two poses.
The number of interpolated frames is decided by the distance between the two frames. 
%We illustrate the interpolated poses in  
%Fig.~\ref{fig:pose_interpolation}. 
It can be observed in Fig.~\ref{fig:pose_interpolation} that under this strategy, the transition between the two original poses remains smooth, whereas the interpolated poses appear as a person performing atypical actions. All the original and interpolated action sequences are used to render RGB action videos in our data generation scheme.

\begin{figure}[t]
\centering
\includegraphics[width=0.48\textwidth]{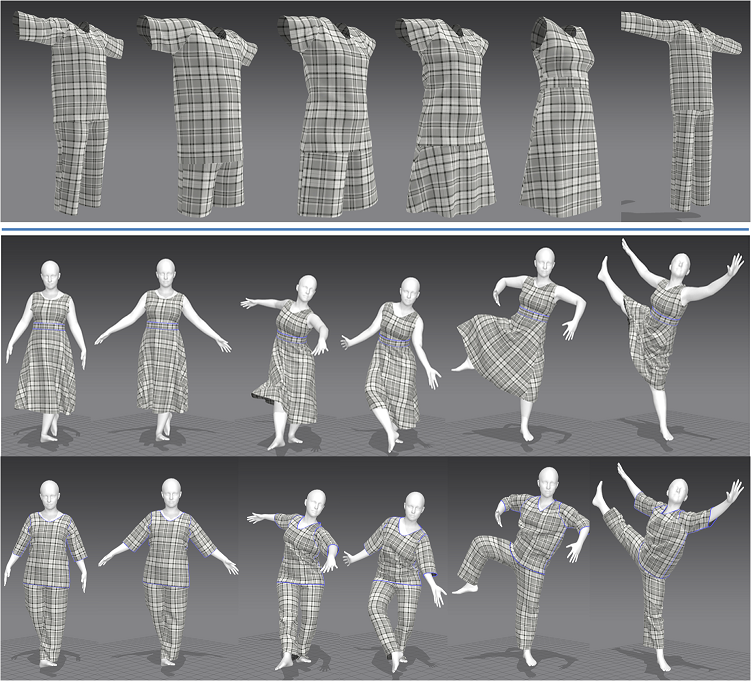}
\caption{First row: Types of garments designed in this work to apply to SMPL models. Second and third row: Physics based application of garments to sequence of poses performing dance moves.}
\label{fig:garment_simulation}
\vspace{-3mm}
\end{figure}

\subsection{RGB Actions with Realistic Clothes}
\label{sec:MD_Cloth}
The avatars resulting from SMPL representation can only represent humans with minimal or tightly-fitted clothes. Consequently, previous attempts of using this representation to generate data in the broader domain of human actions, e.g.~\cite{varol2017learning} extracted (unwrapped) texture maps from human scans, and fitted those onto the avatars. On one hand, the texture variations become limited under this strategy. On the other, this scheme does not correctly imitate clothes in the real-world. 
For instance, simple texture on an avatar does not behave in a `cloth-like' manner, even for slightly loose clothes. Moreover, it does not cause any occlusion to the body shape, which is often there for the loose clothes. Not to mention, using only the texture also deprives the generated data of important temporal cues that cloth dynamics provide in the real-world actions.

\begin{figure}[t]
\centering
\includegraphics[width=0.48\textwidth]{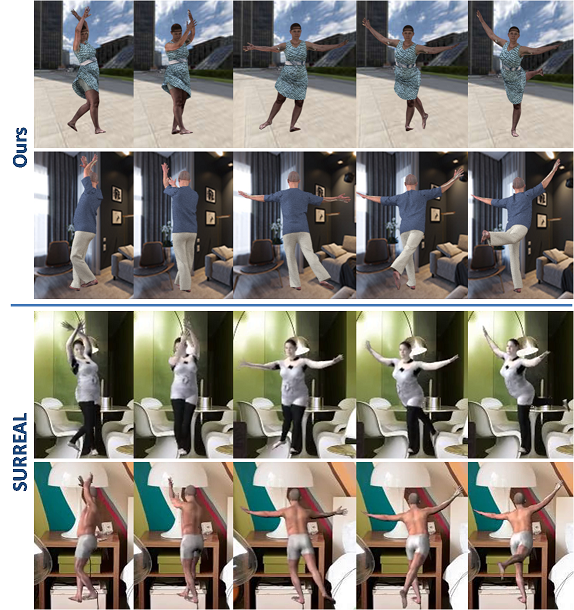}
\caption{Comparison of an action sequence rendered by our method and SURREAL~\protect\cite{varol2017learning}. Our method enables fine cloth details and realistic cloth-body interactions, while SURREAL fits unwrapped textures onto the avatars,  resulting in a less realistic ``body-painting'' effect. See the top of female affixed to legs in the bottom row.}
\label{fig:SURREAL_comparison}
%\vspace{-3mm}
\end{figure}

We address this problem by modeling a number of garments for the SMPL avatars, and using those for data generation.
In this work, the garment simulation is performed with a soft-body Physics engine. % that  supports cloth pattern cutting, sewing and its physical simulation. 
Such engines are used by fashion designing software, e.g.~MarvelousDesigner7 (MD7) to achieve realistic effects. We adapt the engine from MD7 that supports cloth pattern cutting, sewing and its physical simulation.  
In the first row of Fig.~\ref{fig:garment_simulation}, we illustrate different types of garment designed in this work. The second and third rows of the figure illustrate the results of applying two of these garments to a SMPL model under different poses of the avatar with the Physics engine based simulation.
The shown six poses are sampled from a sequence (left to right) of a female avatar performing dance moves.
Notice the realistic effects in terms of wrinkles, draping and the overall cloth movement in our data. 
We show a single texture design for all the avatars for better visualization. 
It is apparent that, applying static textures on avatars simply can not provide the fine details and the realistic interactions between cloth and human body that is provided by our technique.  
We also provide a qualitative comparison of the color frames constructed with our data generation technique with a popular existing method that uses texture-based clothing in Fig.~\ref{fig:SURREAL_comparison}.

%Comparison of rendered images in Fig.~\ref{fig:SURREAL_comparison} also demonstrates this difference

We simulate cloth movements with the help of time-varying partial differential equations, which are solved as ordinary differential equations employing discritization~\cite{baraff1998large}. In our case, a cloth is modelled as a set of particles $[m_i, x_i]$ with interconnecting springs, where $m_i$ and $x_i$ are mass and geometric state vector of the $i$-th particle. The dynamics of the cloth is governed by the Newton`s Equation:
\begin{equation}
    m_i\ddot{x_i} =  F_{int} + F_{ext},
\end{equation}
where $F_{int}$ and $F_{ext}$ respectively denote the collective internal and external forces acting on the cloth, and $\ddot{x_i}$ is the second derivative of $x_i$ with respect to the time.  
Identifying the internal forces in cloth deformation is a hard problem. However, it  can be reduced to the problem of  differentiating the potential energy $E$ of the cloth particles spatially. This simplification leads to the following relationship:
\begin{equation}
   - \frac{\partial E}{\partial x} = F_{int}.
\end{equation}
The equation governing internal forces on a  particle $x_i$ hence becomes,
\begin{equation}
    m_i\ddot{x_i} = -[\frac{\partial E}{\partial x}]_{x=x_i}.
\end{equation}
Putting the cloth particles into vectors, and accounting for the overall external forces, we can re-write our equation as:
\begin{equation}
    M\ddot{x} = -\frac{\partial E}{\partial x} + F_{ext},
\end{equation}
where M is a diagonal matrix formed with  $m_i$, representing mass distribution of the cloth, and $F_{ext}$ models the extra external forces acting on the cloth as air-drag, contact and constraint forces, internal damping, etc. The overall external force on a cloth is also a function of $x$ as well as $\dot{x}$. Taking the time into account, the above equation can be written as
\begin{equation}
    \ddot{x}(t) = M^{-1}(-\frac{\partial{E}}{\partial{x}} + F)(t).
\end{equation}
Our Physics engine solves the above equation numerically~\cite{hauth2005numerical} to get the time derivatives of the cloth particles. Using those with the particle locations results in realistic cloth deformations with its movements. 
The engine computes the Energy $E$ while accounting for the deformations of  `stretch', `shear' and `bending' for the realistic effects.
%These equations account for multiple factors, that include cloth mass distribution, air drag and dampening of the movement etc.
%For instance, one of the equations used by the used  Physics engine is:
%\begin{equation}
%    \ddot{x} = H^{-1}(-\frac{\partial{E}}{\partial{t}} + F).
%\end{equation}
%Here,  the vector $\ddot{x}$ and  matrix $H$ represent the geometric state and mass distribution of the cloth, $E$ yields the cloth’s internal energy, and $F$ model other forces (air-drag, contact and constraint forces, internal damping, etc.) acting on the cloth.
Employing a Physics engine to simulate realistic cloth movements in 3D human videos is a unique contribution of this work.  Our implementation and resulting data will be made public for the broader research community.

\subsection{Scene Variations in Videos}
\label{sec:Scene_Variation}
We generate action videos that are rich in scene variations. We apply different backgrounds, cloth textures, illuminations, and viewpoints to render SMPL avatars to actual RGB frames.
Below, we describe these variations and our method to incorporate them in more detail.

%We maximize scene variations of synthetic dataset by applying different backgrounds/textures, illuminations, and viewpoints during the rendering from SMPL avatars to actual RGB frames. In this section, we explain approaches to maximize the dataset variation, and in next section, we will describe the rendering steps.
%By maximizing the scene variation, the neural network is expected to learn a pose recovery model that is able to eliminate all changing factors of the scene, and therefore focus on the human poses learning to gain better generalization capacity.
\vspace{2mm}
\noindent{\bf Background and cloth 
texture variations:~}
We use over 400 spherical images for the environmental backgrounds in our videos.
These images are collected from online resources, such as Google Images. Our video generation pipeline additionally employs conventional 2D images for  background during the rendering process. Towards that end, we exploit the Places365-Standard dataset~\cite{zhou2018places} and use its test split to generate different backgrounds. 
We set the scales and rotations of the background scenes at random during the rendering process. 
To vary the cloth textures, we use  the DTD~\cite{cimpoi14describing} and Fabrics~\cite{kampouris2016fine} datasets and randomly choose a texture for each video.

%We downloaded over 400 spherical background images from Google Images, and apply them as environmental background. In addition to the spherical background, we also apply normal 2D background images during the rendering. We download Places365-Standard~\cite{zhou2018places} dataset, and use its test split which includes more than 300,000 images of different scenes. This significantly boosts the background variation of our generated synthetic dataset. The scale and rotation of scene background are randomly set during the rendering.

%We increase the cloth variation by applying various textures for the designed garments. DTD~\cite{cimpoi14describing} and Fabrics Dataset~\cite{kampouris2016fine} are downloaded for this purpose.

\vspace{2mm}
\noindent{\bf Lighting variations:~}
We setup four surface lights pointing towards the SMPL avatars and randomly change their strengths during rendering to simulate illumination variations encountered in the real-world scenes. 

\vspace{2mm}
\noindent{\bf Viewpoint variations:~}
We render the moving SMPL avatar from four camera viewpoints setup in the East, West, South, and North of the avatar. To keep  the avatars in the frame center, we set the cameras to track the \textit{Pelvis} joint of the SMPL model. %To offset the relative movement between object and camera, we set camera's focal length as 180mm and shoot the object from a long distance, so that the scale of human model appeared in rendered images will keep the same. 
The camera sensor size and the focal length are set to 32mm and 180mm respectively, and the output resolution is fixed to 250 $\times$ 250. Since we use a telephoto camera, we carefully scale the background during the image rendering to better blend the rendered human models and their backgrounds. Different hyper-parameters of the setup discussed above are optimized empirically to achieve the best realistic visual appearances in the resulting videos.  

%\subsubsection{Occlusion}
%We create several simple 3D objects and randomly place them between the camera and human model.

\begin{figure*}[t]
\centering
%\begin{minipage}[b]{\linewidth}
\includegraphics[width=1\textwidth]{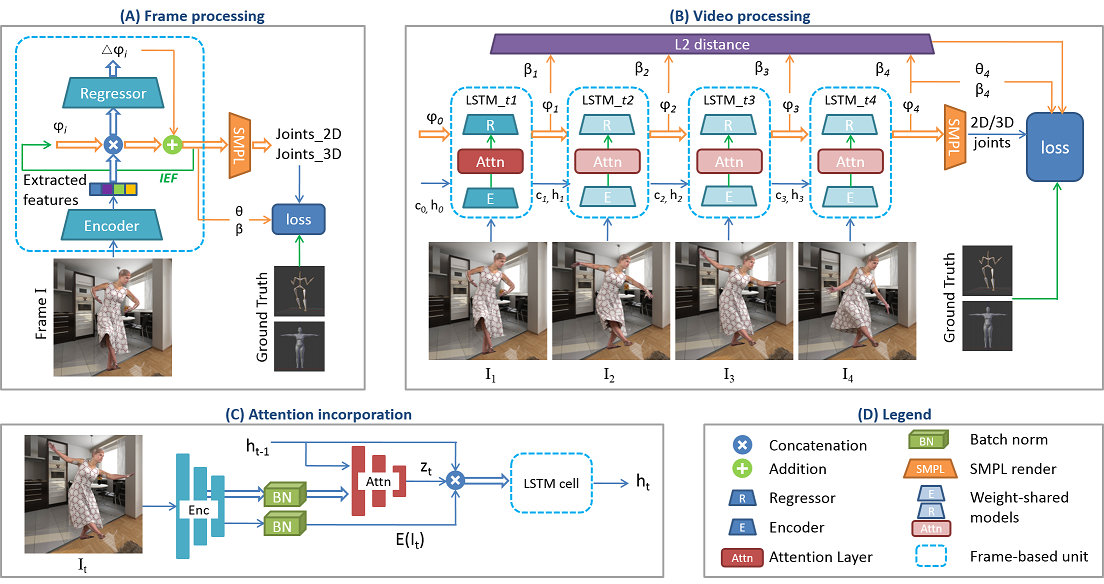}
\caption{Approach schematics: (A) Basic building block to process individual frames. A frame ${\bf I}$ is encoded and a Regressor is trained using Iterative Error Feedback (IEF). Rich 3D annotations provide ground truth for SMPL parameters $\boldsymbol\beta, \boldsymbol{\theta}$ and 2D, 3D skeletons. (B) The block is used in a recurrent setup with as many LSTM cells as the clip length that share Encoder and Regressor weights. An additional collective loss is defined over $\boldsymbol{\beta}_t$ for geometric consistency across the frames. The LSTM cells implicitly model temporal dynamics of a  video. (C) Shows the attention mechanism  that is incorporated to encode contextual information from low-level visual features to condition refreshing of the recurrent states. (D) Legend for the figure.}
\label{fig:video_framework}
%\vspace{-3mm}
\end{figure*}

%\vspace{-2mm}
\subsection{Synthesizing the Videos}
Our data generation pipeline has three main steps to get from the MoShed CMU action sequences to realistic videos. In the first step, we import the SMPL avatar and its respective  MoShed CMU sequences into Blender - an open source 3D  rendering software (\url{https://www.blender.org/}) - and 3D render them.  We achieve 3D avatar animation in this step with bodies that do not have clothes. In the second step, we apply clothes and perform Physics based simulation of the clothed avatars and record their mesh information. The physical properties of the clothes are preset to emulate realistic cloth-body and cloth-gravity interactions in this phase. In the third step, we again use the Blender with  clothed meshes and synthesize videos by varying different scene attributes as discussed in \S \ref{sec:Scene_Variation}. 
%In Figure~\ref{fig:render_samples} we show six different groups of rendered samples. In each group, the realistic cloth deformation and cloth-body interactions are magnified.

Our cloth simulation requires human models in animation to start from a standard ``T-shape'' or ``A-shape'' pose, which are not guaranteed for many actions in CMU MoCap. To address this issue, we interpolate extra $N$ frames prior to the target MoCap action sequence, to ensure a smooth transition from a standard ``T-shape''pose to the real poses of interest.
% This transition is demonstrated in Figure~\ref{}.
In this work, we render videos for more than $5,000$  sequences using four different viewpoints. In total, we generate more than 3 million video frames in our dataset that can find applications in training deep models for various tasks, e.g.~human action recognition, pose estimation.

\section{Full Mesh Pose Recovery}
\label{sec:RecovModel}
The proposed data generation technique enables effective full mesh human pose sequence recovery   directly from a video. To achieve this, we design a recovery mechanism for individual video frames and apply it in a recurrent structure for videos, as illustrated in Fig.~\ref{fig:video_framework}. 

%Our SMPL-based synthetic action video dataset opens the door for learning a video-based 3D human pose recovery model. In this section, we explain our proposed video-based 3D pose recovery framework. The general idea is to first define a image-based building block, and then embed it into a recurrent structure, such that the resulting model is ``video-based'' and able to process continuous frames.

\subsection{Mesh Recovery from a Frame}
\label{sec:frame}
Our data generation technique provides realistic video frames for which we also have rich 3D annotations. We use this data to learn a neural network model that can predict a 3D body mesh with accurate body shape and pose in individual frames.
Concretely, this model predicts a vector $\boldsymbol{\varphi}=[\boldsymbol\theta, \boldsymbol\beta, {\bf R}, {\bf t}, s]$, where $\boldsymbol\theta$ and $\boldsymbol\beta$ are SMPL parameters (see \S \ref{sec:SMPL}), ${\bf R} \in \mathbb R^{3 \times 3}$ describes the global rotation, ${\bf t} \in \mathbb R^2$ records  the translation within the frame and $s \in \mathbb{R}$ is  the scale of the 3D mesh. As shown in Fig.~\ref{fig:video_framework}(A), there are two main components  of the model, an `Encoder' and a `Regressor'.
We implement the Encoder as a CNN and the Regressor as an MLP. We defer the exact  implementation details of these networks to \S \ref{sec:Impl} for continuity.

For an input video frame ${\bf I}$, the Encoder computes a feature vector $\mathcal E(I) \in \mathbb{R}^D$, which is regressed by the Regressor to estimate $\boldsymbol\varphi = \mathcal R(\mathcal E({\bf I}))$. To keep our model compact, we adopt Iterative Error Feedback (IEF) to train the Regressor. In IEF, the $i^{\text{th}}$ iteration computes $\boldsymbol\varphi_{i+1}=\boldsymbol\varphi_i + \Delta\boldsymbol\varphi_i$,  where $\Delta\boldsymbol\varphi$ is the increment. This procedure is initialized with the $\boldsymbol\varphi$ resulting from the mean pose in our dataset.
The predicted vector $\boldsymbol{\varphi}$ can be used to generate a 3D pose mesh ${\bf P} \in \mathbb{R}^{3 \times N}$ under the SMPL representation. Moreover, this mesh can be regressed to achieve 3D skeleton~\cite{loper2015smpl} that we denote by $_{3D}\hat{{\bf J}}\in \mathbb{R}^{3 \times Q}$, where $Q$ is the number of body joints. We further project the 3D joints to 2D key points as follows
\begin{equation}
    _{2D}\hat{{\bf J}} = s\boldsymbol\Psi({\bf R}~~ _{3D}\hat{{\bf J}} ) + {\bf t},
\end{equation}
where $\boldsymbol\Psi(.)$ is the orthographic projection operator. We use the 2D skeleton joints to define a `projection loss' 
\begin{equation}
    \mathcal L_{proj} = \sum\limits_{i } ||\chi_i (_{2D}{\bf J}_i - _{2D}\hat{{\bf J}}_i)||_1,
\end{equation}
where $\chi(.)$ is a masking operator that turns those joints to zero that are not visible in the ground truth 2D skeleton, and $||.||_1$ computes the $\ell_1$-norm. On similar lines, we also define a 3D joint loss  as follows
\begin{equation}
    \mathcal L_{3Djoint} = \sum\limits_{i} || _{3D}{\bf J}_i - _{3D}\hat{{\bf J}}_i||_2^2.
\end{equation}
Moreover, we also define a SMPL parameter loss as
\begin{align}
    \mathcal L_{smpl} = \sum\limits_{i} ||[\boldsymbol\beta_i, \boldsymbol\theta_i] - [\hat{\boldsymbol\beta}_i, \hat{\boldsymbol\theta}_i]||_2^2.
\end{align}

Finally, the overall loss for our network is defined as a combination of the above described losses, given as 
\begin{equation}
   \mathcal L = \mathcal L_{proj} + \delta( \mathcal L_{3Djoint} + \mathcal L_{smpl}).
\end{equation}
In this formulation, the parameter $\delta$ is set to 1 when the ground truth 3D annotation is available and 0 otherwise. %{\color{red} A sentence on why/when these cases occur!!}. 
This parameter is useful when a training batch includes samples with only 2D keypoints annotations. 
Human pose recovery by Kanazawa~\textit{et al.}~\cite{kanazawa2018end} can be related to our technique for frame based mesh recovery. 
%Our network for mesh recovery from a frame is inspired by~. 
However, our full mesh recovery method goes beyond individual frames to videos, as described below.

\subsection{Mesh Recovery from a Video}
\label{sec:videoRe}
Our data allows full mesh based  supervised learning of deep models directly from videos, which was previously not possible.
To recover meshes from videos,  we treat our mechanism for frame processing (\S~\ref{sec:frame}) as a basic building block and use it in a recurrent structure. We also incorporate attention layer in this recurrent model. The attention layer aims to encode low-level visual features of input images, and provide contextual conditions on the  recurrent operations.
The resulting technique explicitly models the temporal dynamics of action sequences and is  able to enforce geometric consistency on the reconstructed poses across the frames. This makes the recovered 3D poses appear more  natural and realistic, as illustrated in  Fig.~\ref{fig:teaser}.

%Direct 3D supervision effectively mitigates the depth ambiguity problem in image-based pose recovery task, and it is also critical in extending the recovery task from image-based to video-based. 

%Direct 3D supervision in pose recovery task requires large corpus of ``in-the-wild'' RGB images with annotated 3D pose data, which is currently not available. Although some method works around the lack of such data by learning adversarial prior, we believe that direct 3D supervision is more valuable and thus can not be replaced. The reason is that the prior learned from unrelated pose data suffers from possible data distribution bias, and moreover, incorporating adversarial regularization in temporal system can significantly increase the model complexity and thus degenerate the learning performance. As a result, our proposed novel video-based recovery framework is built based on the synthetic action videos, which provides direct and accurate 3D supervision for the learning of 3D pose recovery task.

%As mentioned before, we take the image-based recovery model~\ref{sec:HMR} as a basic building block to create our video-based 3D pose recovery model. The video-based model takes advantage of our synthetic ``in-the-wild'' action videos, and explicitly models the temporal variation of action sequence. Moreover, the video-based model enforces the geometric consistency of reconstructed poses along temporal dimension, resulting in a natural and real recovered 3D pose sequence.

The proposed network for full mesh pose recovery is illustrated in Fig.~\ref{fig:video_framework}~(B).
We embed the frame-based building block in a recurrent structure by  sharing the weights of the Encoder and Regressor. 
Given an input video clip $\{{\bf I}_t\}$, s.t.~$t\in\{1,2,...,T\}$; where $T$ is the number of frames in the clip,  the same Encoder is used to extract visual features from each frame. 
In addition to the feature vector $\mathcal E(I_t)$, we probe the low-level feature map $\mathcal E(I_t)_{low} \in \mathbb{R}^{L \times D}$, which can be represented as a set of annotation vectors $a_t=\{a_t^1,...,a_t^L\}$, $a_t^i\in\mathbb{R}^D$. With an attention layer $f_{att}$ which is implemented as a multi-layer perceptron (MLP)~\cite{xu2015show}, the context vector $\hat{z}_t$ of frame $\bf I_t$ is encoded with the annotation vectors $a_t$.
\begin{equation}
\label{func:attn_1}
    e_t^i = f_{att}(a_t^i),
\end{equation}
\begin{equation}
    \alpha_t^i = \frac{\mathrm{exp}(e_t^i)}{\sum_{k=1}^{L}\mathrm{exp}(e_t^k)}.
\end{equation}
Here, $\alpha_t^i$ represents the weights of each annotation vector $a_t^i$ and $\sum_{i=1}^{L}\alpha_t^i=1$. Intuitively, it represents the importance of particular visual feature elements, and, therefore, indicates ``where" and ``how much" the network should pay attention to. With the calculated weights $\alpha_t^i$, we adopt a deterministic ``soft'' attention mechanism~\cite{bahdanau2014neural} to compute the context vector $\hat{z}_t$ as follows
\begin{equation}
    \hat{z}_t = \sum_{i=1}^{L}{\alpha_t^i a_t^i}.
\end{equation}

For the recurrent structure, we employ LSTM~\cite{hochreiter1997long} for its gate and memory design that makes its training more effective. 
Following the implementation in~\cite{zaremba2014recurrent}, the overall behaviour of LSTM is represented mathematically by
\begin{equation}
    c_t = f_t \odot c_{t-1} + i_t \odot g_t,
\end{equation}
\begin{equation}
    h_t = o_t \odot \text{tanh}(c_t).
\end{equation}
In the above equations, $i_t,f_t,c_t,o_t,h_t$ are the input, forget, memory, output and hidden state of the LSTM at time step $t$, and $\odot$ denotes the Hadamard product. Based on the above equations, we denote the LSTM cell function as $h_t=\Omega(i_t)$ which controls the inflow and outflow of information through the LSTM at $t^{\text{th}}$ frame. The internal states of LSTM are controlled by the affine transformations with trainable parameters.
The total time step of LSTM is set to $T$. %the video length in a training sample, i.e.~$T$.

To incorporate the attention mechanism into the recurrent network, we make the context vector $\hat{z}_t$ conditioned on the previous hidden state of LSTM, i.e.~$h_{t-1}$. Intuitively, as LSTM advances in a frame sequence, how the network pays its attention is conditioned on the information it has processed in the previous time steps. Hence, we can rewrite Eq.~(\ref{func:attn_1}) as
\begin{equation}
    e_t^i = f_{att}(a_t^i, h_{t-1}).
\end{equation}
Using the above configuration, the network computes the  desired vector $\boldsymbol{\varphi}_t$ for the $t^{\text{th}}$ frame as
\begin{equation}
    \boldsymbol{\varphi}_t = \mathcal R(\Omega(\mathcal E({\bf I}_t)\otimes \hat{z}_t \otimes   \boldsymbol{\varphi}_{t-1}));~ t=1,2,...,T,
\end{equation}
%where $\Omega(.)$ denotes the LSTM cell function that controls the inflow and outflow of information through LSTM's gate and memory, and $\otimes$ is vector concatenation. 
where $\otimes$ denotes vector concatenation.
Note that, the input to an LSTM cell at each time step corresponds to the frame's encoded feature augmented with the contextual information derived from low-level visual features, and the prediction from the last time step. At time 0, we assign the prediction $\boldsymbol{\varphi}_0$ as the mean value of SMPL parameters for our data. As for the cell state $ c_t$ and the output state $h_t$, we initialize them as $(c_0, h_0) = \mathcal{I}(\mathcal E({\bf I}_1))$, where $\mathcal I(.)$ is implemented with an MLP with two hidden layers that maps an encoded feature vector  to the initial states.
This initialization strategy helps in faster training of the overall recurrent network.

To enforce geometric consistency along a predicted pose sequence, we propose an additional shape smoothness loss $\mathcal L_{shape}$, for the recurrent network as
\begin{equation}
    \mathcal L_{shape} = \sum_{t=1}^{T-1}||\boldsymbol\beta_{t+1} - \boldsymbol\beta_t||_2^2,
\end{equation}
where $\boldsymbol\beta_t$ is obtained from the Regressor at the $t^{\text{th}}$ time stamp. %The loss $\mathcal L_{shape}$ is aimed at making the body shape between consecutive frames consistent.
In our recurrent structure, this loss inherently accounts for all the frames in a video clip.
According to~\cite{wei2016convolutional}, intermediate supervision helps in mitigating the vanishing gradient  problem in the recurrent  networks, and also helps in better conditioning of the learning process. 
Hence, we eventually define our overall loss function as
\begin{align}
    \mathcal L = \sum_{t=1}^{T}\lambda((\mathcal L_{proj})_t + \delta \mathcal (L_{3D})_t) + \mathcal L_{shape},
\end{align} 
 where $\mathcal L_{3D}\!\!=\!\! \mathcal L_{3Djoint} + \mathcal L_{smpl}$, and   $\lambda$ is a  hyper-parameter. 

\begin{table*}[t]
\centering
{\footnotesize
\caption{3D joint evaluation results on Human3.6M datasets: Evaluation metric PA-MPJPE, units mm. The proposed method is abbreviated as Mesh VIdeo PosE Recovery (MVIPER). Per action and mean error is reported for two evaluation protocols. See text for protocol description.}
%\vspace{-1mm}
\label{tab:human36m_p}
\setlength{\tabcolsep}{2.5pt}
\begin{tabular}{l|ccccccccccccccc|c}
\hline\noalign{\smallskip}
Protocol-1 & Direc. & Discu. & Eat & Greet & Phone & Photo & Pose & Purchase & Sit & SitD. & Smoke & Wait & WalkD. & Walk & WalkT. & Mean \\
\noalign{\smallskip}\hline\noalign{\smallskip}
%SMPLify & 62.0 & 60.2 & 67.8 & 76.5 & 92.1 & 77.0 & 73.0 & 75.3 & 100.3 & 137.3 & 83.4 & 77.3 & 79.7 & 86.8 & 81.7 & 82.3 \\ 
HMR~\cite{kanazawa2018end} & 52.3 & 54.7 & 54.3 & 57.1 & 60.9 & 70.4 & 51.6 & 49.9 & 65.7 & 76.0 & 58.6 & 52.5 & 60.2 & \textbf{45.2} & \textbf{53.6} & 57.5 \\
HMR$\dagger$ & 51.2 & 52.4 & 53.8 & 56.9 & 59.9 & 65.0 & 50.4 & 49.2 & 66.3 & 73.1 & 59.2 & 52.6 & 60.0 & 46.6 & 53.9 & 56.7 \\
\hline
MVIPER (ours) & \textbf{48.1} & \textbf{48.8} & \textbf{49.6} & \textbf{55.3} & \textbf{53.8} & \textbf{63.4} & \textbf{49.4} & \textbf{48.0} & \textbf{58.5} & \textbf{67.4} & \textbf{54.4} & \textbf{52.2} & \textbf{59.3} & 47.3 & 54.3 & \textbf{54.0} \\
%MVIPER (ours) &  &  &  &  &  &  & 49.5 &  &  &  &  & 52.0 &  & 45.1 & 51.0 &  \\
\hline\hline\noalign{\smallskip}
Protocol-2 & Direc. & Discu. & Eat & Greet & Phone & Photo & Pose & Purchase & Sit & SitD. & Smoke & Wait & WalkD. & Walk & WalkT. & Mean \\
\noalign{\smallskip}\hline\noalign{\smallskip}
SMPLify~\cite{bogo2016keep} & 62.0 & 60.2 & 67.8 & 76.5 & 92.1 & 77.0 & 73.0 & 75.3 & 100.3 & 137.3 & 83.4 & 77.3 & 79.7 & 86.8 & 81.7 & 82.3 \\ 
HMR~\cite{kanazawa2018end} & 53.2 & 56.8 & 50.4 & 62.4 & 54.0 & 72.9 & 49.4 & 51.4 & 57.8 & 73.7 & 54.4 & 50.0 & 62.6 & 47.1 & \textbf{55.0} & 56.7 \\
HMR$\dagger$ & 52.1 & 53.9 & 51.4 & 61.1 & 54.4 & 66.1 & 49.6 & 48.7 & 58.3 & 69.9 & 54.6 & \textbf{50.0} & 60.6 & 49.3 & 55.5 & 55.7 \\
\hline
MVIPER (ours) & \textbf{48.0} & \textbf{46.0} & \textbf{46.0} & \textbf{57.1} & \textbf{48.6} & \textbf{61.3} & \textbf{47.7} & \textbf{46.8} & \textbf{54.1}  & \textbf{67.1} & \textbf{48.9} & 50.1 & \textbf{59.1} & \textbf{47.8} & 56.1 & \textbf{52.3} \\
%MVIPER (ours) & 47.8 &  &  & 56.6 & 51.9 & 61.2 & 46.2 &  &  &  & 52.1 & 48.2 &  & 47.3 & 51.9 &  \\
\hline\hline\noalign{\smallskip}
\end{tabular}}
%\vspace{-4mm}
\end{table*}

%\vspace{-1mm}
\subsection{Implementation Details}
\label{sec:Impl}
%In this section, we explain the implementation details of our proposed approach.
To implement the Encoder (E), we use the   ResNet-50 model~\cite{he2016identity} pre-trained on ImageNet~\cite{deng2009imagenet}.  We consider  the convolution activation prior to Softmax as an image feature, and the activation values of the layer ``resnet-v2-50$/$block4'' as the low-level feature map.
We realize the Regressor (R) as a Multiple Layer Perceptron (MLP) with two fully-connected hidden layers, and an  output layer that has the same dimension as vector  $\boldsymbol{\varphi}$. %Iterative Error Feedback(IEF) is applied for the regressor $R$, and the number of iteration is 3. 
Both E and R are shared by every time step for the LSTM, whereas we empirically set the width of the hidden unit for the LSTM to 2048, and use $T =4$.
Our model for videos is trained in two main steps. First, we train a frame-based model and then use it in the second step for video based training. We copy the Encoder and Regressor weights for the recurrent model for initialization, and then train the model with stochastic gradient descent using  Adam optimizer~\cite{kingma2014adam} in an end-to-end manner. We set the learning rate to $10^{-5}$, and batch size to 16. %We denote the Video-based Human Pose Recovery model in the second stage as VHPR. \

%set $T=4$ for total time steps, and $H=2048$ for the width of hidden unit.

%We train our video-based model in 2 steps. In the first step, we re-train the basic frame-based block using trained weights of HMR~\cite{kanazawa2018end}, with the combination of real and synthetic data. 
%Due to the increase in the scale of training dataset, especially more data with 3D ground truth, we expected the encoder $E$ and regressor $R$ in the re-trained model performs better than those in the original HMR model. We denoted retrained block as HMR$_{re}$, and the retrained encoder and regressor and $E_re$ and $R_re$ respectively. %The performance of enhanced HMR will be evaluated in the section~\ref{sec:3D_Joints_Eval}. 
%In the second step, we proceed to train our video-based model. We copy the weights of $E_re$ and $R_re$ to the video-based recurrent model, and then train the model with stochastic gradient descent and Adam optimizer~\cite{kingma2014adam} end-to-end. The learning rate is set to be $1 \times 10^{-5}$, and batch size is 24. We denote the Video-based Human Pose Recovery model in the second stage as VHPR. Note that both real and synthetic data is used to train VHPR. 

To train a model, we also add samples from LSP, LSP-extended~\cite{Johnson11} MPII~\cite{andriluka20142d}, MS COCO~\cite{lin2014microsoft}, Human3.6M~\cite{ionescu2014human3} and MPIINF-3DHP~\cite{mehta2017vnect} to our data, in addition to the proposed data.
%{\color{red} Are these datasets already useful for Video Mesh Recovery?}.
These datasets have been filtered to remove  images that are too small or have less that 6 visible joints. The standard train/test split is used. Datasets such as LSP, COCO and MPII consist of independent frames, for which we replicate the individual frames to form training video clips. 

% We train the model in first stage for 25 epochs, and train the model in second stage for 15 epochs.

%\vspace{-2mm}
\section{Evaluation}
\label{sec:Eval}

%\subsubsection{Synthetic Dataset}

\subsection{Quantitative Evaluation}
\label{sec:QuantiEval}
\paragraph{Metrics}
Evaluation of a full 3D pose recovery method is not straightforward. It requires computing point-to-point errors between predicted and the ground truth `mesh' points. 
For that, point-to-point mesh registration is required which is often not possible with the existing datasets. Consequently, recent works mostly resort to reporting  Mean Per Joint Position Error (MPJPE) for 3D skeletons. For benchmarking, we also adopt MPJPE as one of our evaluation metrics. Nevertheless, since our data provides the possibility of point-to-point registration and mesh recovery directly from videos,  we further introduce the following new  metrics to more appropriately evaluate a model for video-based full mesh pose recovery: (a)~Mean Per Vertex Position Error (MPVPE), (b)~Mean Running Vertex Position Variation (MRVPV) and (c)~Mean Running Shape Variation (MRSV).

We define the MPVPE as
\begin{equation}%
\text{MPVPE}=\frac{1}{M} \sum_{j=1}^{M}\left(\sum_{i=1}^{N}||{\bf {\hat v}}_i-{\bf {v}}_i||_2\right),
\end{equation}%
where $M$ and $N$ respectively denote the frame length and the vertices in a mesh, and ${\bf v}_i$; and ${\bf \hat{v}}_i$ are the ground-truth and predicted vertex locations, respectively.
This metric can be considered a mesh variant of MPJPE.
To explicitly account for the temporal dimension of a video, we define MRVPV as 
\begin{equation}%
\text{MRVPV}=\frac{1}{M}\sum_{j=1}^{M-1}\left(\sum_{i=1}^{N}||{\bf \hat{v}}_{j+1,i} - {\bf \hat{v}}_{j,i}||_p\right),
\label{func:TVPV}
\end{equation}%
where $||.||_p$ denotes the $\ell_p$-norm. We consider both $\ell_1$ and $\ell_2$ norms in this work, denoting the resulting variants by  MRVPV$_1$ and MRVPV$_2$. 
We also define MRSV as
\begin{equation}
\text{MRSV}=\frac{1}{M}\sum_{i=1}^{M-1}\left(||\hat{\boldsymbol\beta}_{i+1} - \hat{\boldsymbol\beta}_i||_p \right),
\label{func:TBSV}
\end{equation}
denoting its $\ell_1$ and $\ell_2$ norm variants by MRSV$_1$ and MSRV$_2$, respectively. By definition, this metric gives an estimate of how well the shape of the computed avatar is maintained between consecutive video frames. A lower value ensures better geometric consistency in terms of the avatar shape.

\begin{table}[t]
\centering
{\small
\caption{3D Joint evaluation results on the proposed data.}
%\vspace{-1mm}
\label{tab:synth_p1}
\begin{tabular}{lcc}
\hline\noalign{\smallskip}
\multicolumn{ 1}{l}{Method} & \multicolumn{ 1}{c}{MPJPE} & \multicolumn{ 1}{c}{PA-MPJPE} \\ 
\noalign{\smallskip}\hline\noalign{\smallskip}
SMPLify~\cite{bogo2016keep} & 152.1 & 109.3 \\ %\hline
HMR~\cite{kanazawa2018end} & 133.2 & 81.3 \\ %\hline
HMR$\dagger$ & 125.6 & 77.6 \\ %\hline
MVIPER (ours) & \textbf{93.2} & \textbf{60.5} \\ %\hline
\hline\noalign{\smallskip}
\end{tabular}
}
%\vspace{-3mm}
\end{table}

\paragraph{3D Joints Evaluation}
For 3D joint evaluation, we use the standard MPJPE metric. We also follow~\cite{bogo2016keep} to adjust the global misalignment for the reconstructed 3D joints by applying a similarity transform via the Procrustes Analysis~(PA). The adjusted error is then reported as PA-MPJPE.
We use both Human3.6M dataset and the proposed data for evaluation.
Table~\ref{tab:human36m_p} and Table~\ref{tab:synth_p1} summarize our results, where our method is coined as Mesh VIdeo PosE Recovey (MVIPER). % Due to page limit, we report PA-MPJPE only for the Human3.6M, and report both MPJPE and PA-MPJPE for our data. 
In the reported results, HMR$\dagger$ is our enhancement of HMR~\cite{kanazawa2018end}, which is achieved by fine tuning it on our dataset.

To achieve the results in Table~\ref{tab:human36m_p}, we follow the standard practice of using 5 subjects (ID: S1, S5, S6, S7, S8) for training and 2 subjects (ID: S9, S11) for testing. We employ two standard protocols. Protocol-1: uses samples from all four provided viewpoints for testing, and Protocol-2: uses only the frontal viewpoint samples. To evaluate MVIPER under frame based protocols, we replicate a frame multiple times to form a clip. In the Tables, our method is able to outperform the existing methods consistently.
It is also notable that HMR$\dagger$ is able to perform better than HMR demonstrating the effectiveness of the proposed dataset for the task of 3D pose recovery in general. 
We note that this work focuses on recovering full 3D meshes, hence we include only those pose recovery methods in our comparisons that have this ability.

\paragraph{3D Mesh Evaluation}
We evaluate the performance of our method for 3D pose mesh recovery from videos/images, and compare it with the existing SMPL-based methods that have the mesh recovery  ability. 
Table~\ref{tab:synthetic_mesh} summarizes the results of our experiments using the evaluation metrics discussed above.
As can be seen, the proposed method consistently outperforms the state-of-the-art 3D human pose mesh recovery methods. Again, the gain of HMR$\dagger$ over HMR demonstrates the effectiveness of the proposed dataset.

%\vspace{-3mm}

\subsection{Ablation Study}
%One of the major contributions of this work is in the proposal of new dataset with rich 3D annotations. 
The  dataset proposed in this work enables learning 3D pose recovery models with  full supervision in terms of 2D keypoints, 3D skeleton, and SMPL pose and shape parameters.
Our model fully exploits these supervision labels for the pose recovery. In this section, we study the contribution of each of these supervision labels to the overall performance of our technique by re-training the MVIPER model with different loss combinations that are designed to capitalize on different supervision labels. 

\begin{table}[t]
\centering
{\footnotesize
\caption{3D mesh evaluation on proposed dataset.}
%\vspace{-1mm}
\setlength{\tabcolsep}{0.005cm}
\label{tab:synthetic_mesh}
\begin{tabular}{lccccc}
\hline\noalign{\smallskip}
\multicolumn{ 1}{l}{Method} & \multicolumn{ 1}{c}{MPVPE} & \multicolumn{ 1}{c}{MRSV$_1$} & \multicolumn{ 1}{c}{MRSV$_{2}$} & \multicolumn{ 1}{c}{MRVPV$_{1}$} & \multicolumn{ 1}{c}{MRVPV$_{2}$} \\
\noalign{\smallskip}\hline\noalign{\smallskip}
SMPLify~\cite{bogo2016keep} & 1426.9 & 0.85 & 0.41 & 257.9 & 4.64 \\ %\hline
HMR~\cite{kanazawa2018end} & 1056.5 & 0.82 & 0.36 & 194.0 & 4.31 \\ %\hline
HMR$\dagger$ & 923.7 & 0.76 &  0.32 & 191.2 & 4.25 \\ %\hline
MVIPER (ours) & \textbf{692.7} & \textbf{0.51} & \textbf{0.29} & \textbf{178.7} & \textbf{4.02} \\ %\hline
\hline\noalign{\smallskip}
\end{tabular}}
%\vspace{-3mm}
\end{table}

The first column of Table~\ref{tab:ablation_h36m} shows the loss combinations used for this ablation study. For each combination, we re-train the MVIPER model and report quantitative  results on Human3.6M dataset under Protocol-1 and Protocol-2. Both MPJPE and the adjusted error PA-MPJPE are reported. The results clearly demonstrate   the importance of direct 3D supervision for the  pose recovery task. With 2D objective $\mathcal L_{proj}$, the model learning suffers from serious depth ambiguity. With 3D supervision,  in both cases of $\mathcal L_{3Djoint}$ or $\mathcal L_{smpl}$, the accuracy improvement is remarkable. Considering rows 2 and 3 of the table, $\mathcal L_{proj}$+$\mathcal L_{3Djoint}$ achieves higher accuracy in MPJPE than $\mathcal L_{proj}$+$\mathcal L_{smpl}$, while both combinations have comparable performance with global alignment (PA-MPJPE). This is intuitive because both $\mathcal L_{3Djoint}$ and $\mathcal L_{smpl}$ reflect full 3D joints information, while $\mathcal L_{3Djoint}$ emphasizes more on the absolute joint locations, which benefits the 3D joints evaluation. The last row of Table~\ref{tab:ablation_h36m} demonstrates that MVIPER achieves the best performance when we  combine all the loss terms.

\begin{table}[t]
\centering
{\footnotesize
\caption{Ablation study for training MVIPER with different loss combinations. For each row, MVIPER is re-trained with the designated losses and then evaluated on Human3.6M dataset. Definition of losses can be found in Section~\ref{sec:frame}.}
%\vspace{-1mm}
\label{tab:ablation_h36m}
\setlength{\tabcolsep}{0.005cm}
\begin{tabular}{l|cc|cc}
\hline\noalign{\smallskip}
\multicolumn{ 1}{l|}{} & \multicolumn{ 2}{c|}{Protocol-1} &\multicolumn{ 2}{c}{Protocol-2} \\ 
\noalign{\smallskip}\hline\noalign{\smallskip}
Loss Combinations & MPJPE & PA-MPJPE & MPJPE & PA-MPJPE \\ 
\noalign{\smallskip}\hline\noalign{\smallskip}
$\mathcal L_{proj}$ & 139.0 & 72.6 & 117.3 & 67.2 \\ %\hline
$\mathcal L_{proj}$+$\mathcal L_{3Djoint}$ & 82.6 & 56.4 & 83.7 & 55.7 \\ %\hline
$\mathcal L_{proj}$+$\mathcal L_{smpl}$ & 110.7 & 58.2 & 105.3 & 57.6 \\ %\hline
$\mathcal L_{3Djoint}$+$\mathcal L_{smpl}$ & 83.8 & 58.0 & 88.9 & 58.3 \\ %\hline
$\mathcal L_{proj}$+$\mathcal L_{3Djoint}$+$\mathcal L_{smpl}$ & \textbf{82.2} & \textbf{54.0} & \textbf{81.5} & \textbf{52.3} \\ %\hline
\hline\noalign{\smallskip}
\end{tabular}}
%\vspace{-3mm}
\end{table}

\subsection{Qualitative Evaluation}
\label{sec:QualiEval}

\begin{figure}[t]
\centering
\includegraphics[width=0.48\textwidth]{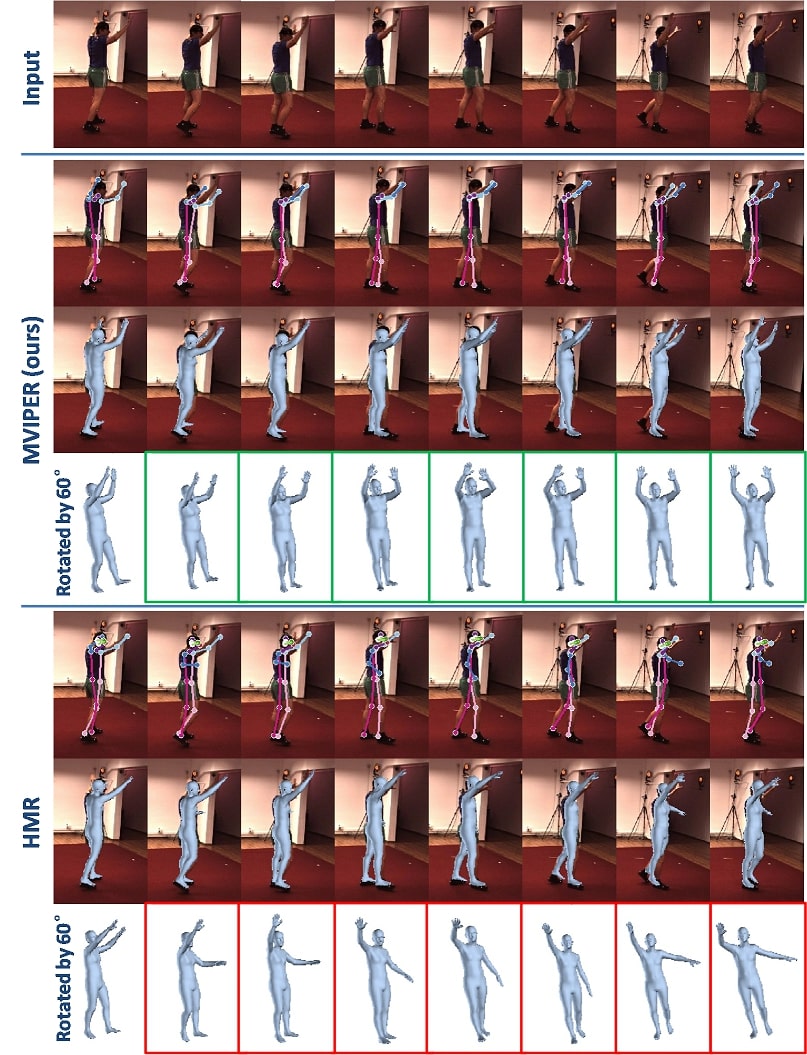}
\caption{Qualitative comparison for HMR and our approach on Human3.6M. For each method, the recovered mesh and its 60$^\circ$ rotation are displayed. The rotated meshes are shown for better visualization.}
\label{fig:qual_comp}
\end{figure}

\begin{figure}[t]
\centering
\includegraphics[width=0.48\textwidth]{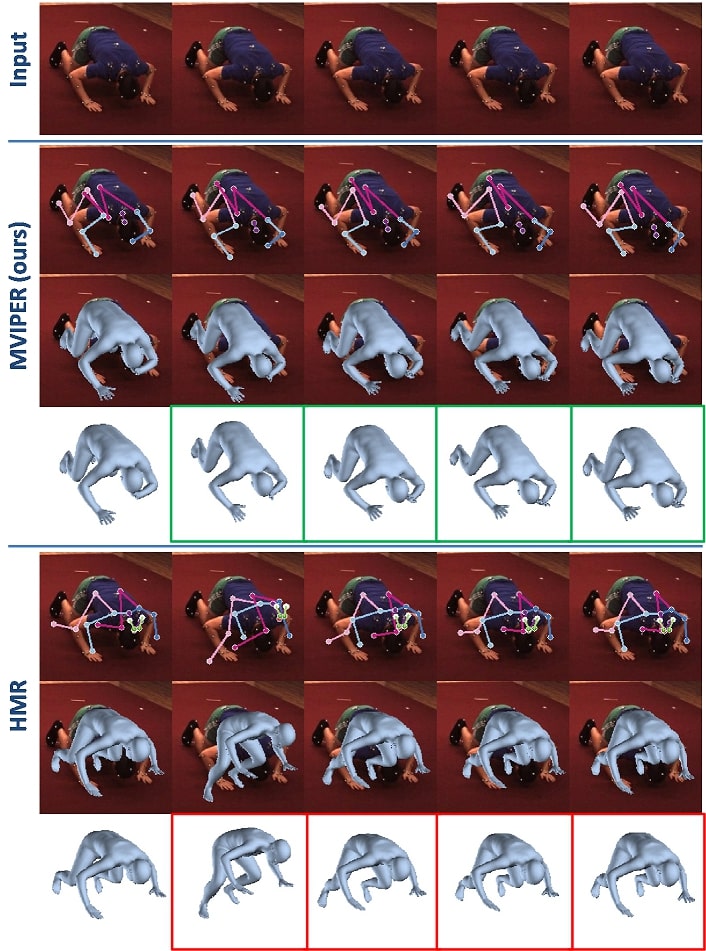}
\caption{Qualitative comparison for HMR and our approach on Human3.6M. Our approach recovers more accurate and coherent pose sequence.}
\label{fig:qual_comp_0}
%\vspace{-3mm}
\end{figure}

\begin{figure}[t]
\centering
\includegraphics[width=0.48\textwidth]{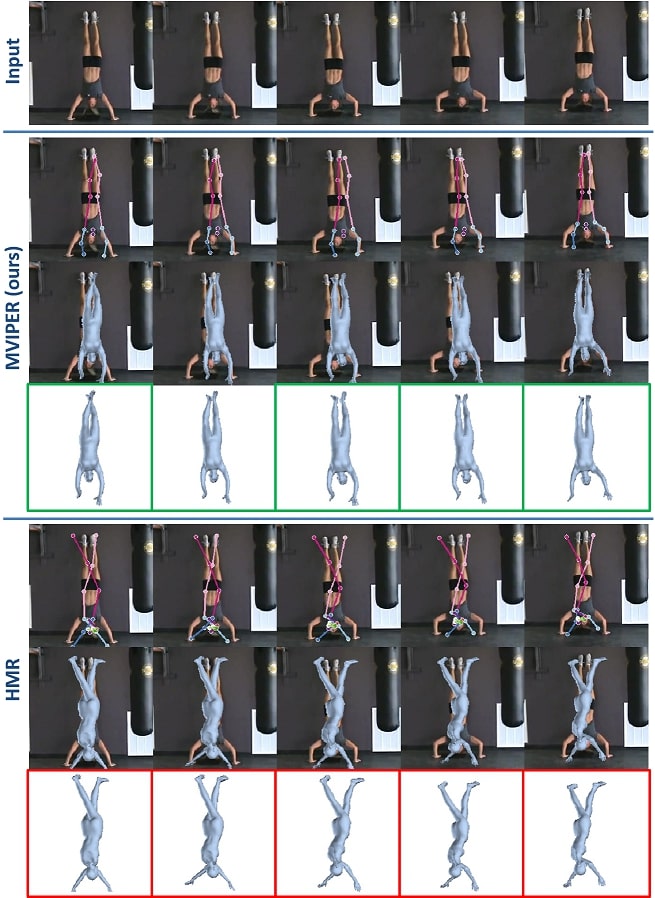}
\caption{Qualitative comparison. Challenging ``handstand'' poses are accurately recovered by our method because it benefits from a large amount of pose variations that are present in the proposed dataset.}
\label{fig:qual_comp_2}
%\vspace{-3mm}
\end{figure}

\begin{figure}[t]
\centering
\includegraphics[width=0.48\textwidth]{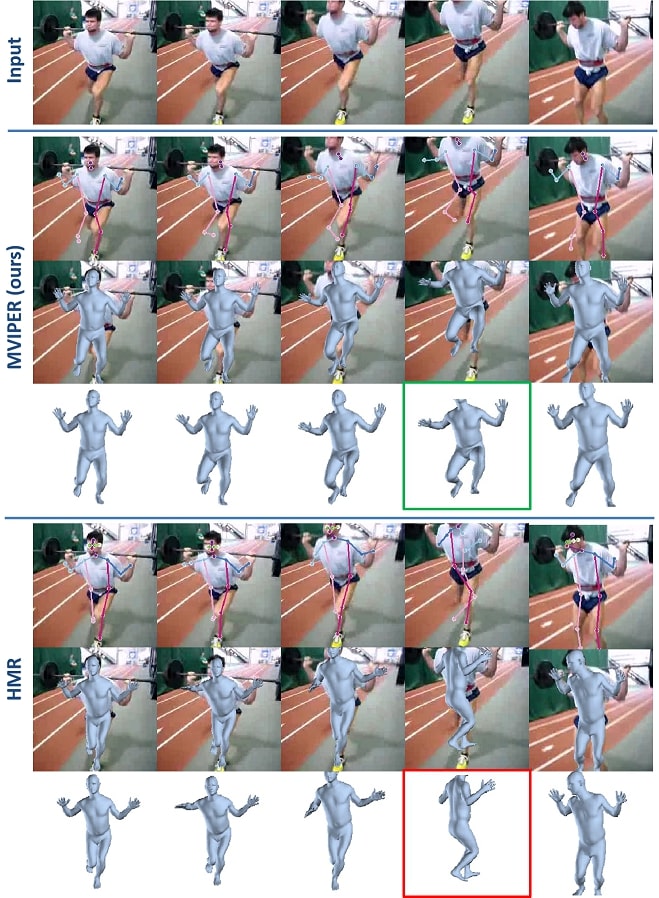}
\caption{Qualitative comparison. HMR suffers from a sudden body orientation error, as it does not consider the temporal context from neighbouring frames.}
\label{fig:qual_comp_3}
%\vspace{-3mm}
\end{figure}

\begin{figure}[t]
\centering
\includegraphics[width=0.48\textwidth]{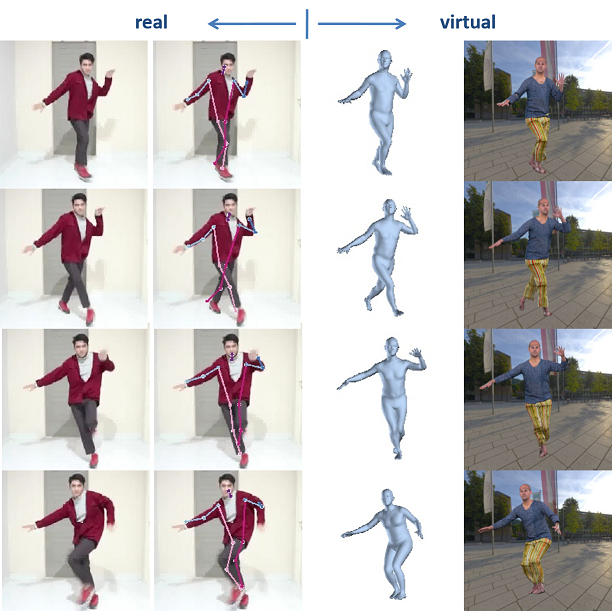}
\caption{Example of motion transfer performed using the proposed MVIPER.  Motions are transferred from real to virtual world with new cloth and background in the virtual world.}
\label{fig:appl_1}
\end{figure}

Representative examples for qualitative performance comparison between our method and the competing frame-based technique, i.e.~HMR is shown in Fig.~\ref{fig:qual_comp} and Fig.~\ref{fig:qual_comp_0}. The examples are  selected at random from Human3.6M dataset. Note that, Human3.6M dataset does not provide ground truth for 3D meshes, hence we can only show the qualitative results. In Fig.~\ref{fig:qual_comp},  when the performer's left arm is occluded by  body, HMR loses its spatial cues and fails to reconstruct it appropriately, while our approach is able to construct the left arm properly by accounting for the temporal cues. The 60$^\circ$ rotation of mesh is provided for better visualization. Similarly, for the complex motions, as illustrated in  Fig.~\ref{fig:qual_comp_0}, HMR also fails to recover accurate and consistent poses. On the other hand, the proposed method performs a much more accurate pose recovery.

 In addition to Human3.6M, we further conducted qualitative analysis on UCF101~\cite{soomro2012ucf101} - an ``in-the-wild'' action dataset unseen by our method and the benchmark HMR method during training. From the UCF101 dataset, we selected  action sequences in which body motions represent the major contents of video frames. 
We show representative qualitative results on this dataset in  %Fig.~\ref{fig:qual_comp_1},
 Fig.~\ref{fig:qual_comp_2}. 
and Fig.~\ref{fig:qual_comp_3}. 
For each sequence evaluated by MVIPER and HMR, we highlight the frames which show significant quality difference. Our method demonstrates consistent advantage over frame-based HMR. In Fig.~\ref{fig:qual_comp_2}, HMR fails to recover the uncommon ``handstand'' poses, while our method recovers it properly. We attribute this difference to our synthetic training data which includes more pose variations, and hence improves the performance of the resulting model. As our method models  temporal variations and imposes geometric consistency, it is able to generate smooth transition for recovered pose sequences. On the other hand, frame-based methods like HMR suffer from temporal jitters and sudden orientation error as visible in  Fig.~\ref{fig:qual_comp_3}.

\subsection{Video-based Motion Transfer}
\label{sec:Motrans}
This is the first work to explore video-based full mesh pose recovery. This problem can find many interesting applications in e.g.~animations. As a possible application of our approach we explore video-based motion transfer. In this application, we input a real-world video of a human performing an action and transfer that motion to an avatar in virtual world. The proposed MVIPER uses the input video to compute 3D meshes for each frame, which are subsequently rendered with different cloths and backgrounds to create virtual actions. The whole process transfers a real-life human motion to a virtual world  avatar. We illustrate a representative example from this experiment in Fig.~\ref{fig:appl_1}. As can be seen, the proposed technique is able to transfer the motion with good fidelity due to accurate human mesh recovery.

%As a possible application used in automatic animation or virtual reality, Fig.~\ref{fig:appl_1} shows an example of motion transfer fulfilled with our approach. First, a real action sequence is fed into MVIPER model to recover 3D pose meshes, and then the meshes are rendered with different cloths and backgrounds to create virtual actions. In this process, the motion in source action is retrained and transferred from real to virtual world.

\section{Conclusion}
\label{sec:Concl}
%\vspace{-1mm}
%We proposed a data generation technique and an end-to-end trainable RNN to recover full 3D meshes of human poses directly from videos.
%Our data generation method  exploits Computer Graphics and Physics to generate realistic videos with a large variety of human pose and scene variations. 
%We use the data to train our neural model that uses a basic building block in a recurrent setting, and imposes geometric consistency of meshes across the video frames. We evaluated the proposed method on our dataset and Human3.6M, achieving very promising results for full mesh recovery from  video. 

We proposed a data generation technique and an end-to-end trainable RNN to recover full 3D meshes of human poses directly from monocular videos. Our data generation method exploits Computer Graphics to generate realistic action videos with a large amount of scene variations in  backgrounds, cloth textures, illuminations and viewpoints. To further enhance action and pose variations in the generated data, pose interpolation is employed to create novel pose sequences between largely varied pose pairs. Moreover, we embed a Physics engine in data generation to produces vivid cloth deformations and cloth-body interactions. This is the first successful application of a Physics engine in contemporary human data synthesis technique for learning deep models. By using the proposed action video dataset and a parameterized human model, we also developed a neural network for video-based full mesh pose recovery.
Our network embeds a basic building block in a recurrent structure and explicitly encodes temporal variations of input video frames, and imposes geometric consistency over the recovered meshes across the video frames. We evaluated the proposed method on our dataset and Human3.6M, using conventional per-joint error metrics, as well as more advanced per-vertex error metrics introduced in this work. Qualitative comparison is provided on Human3.6M and UCF101 action datasets. Both quantitative and qualitative results demonstrate that our method achieves very promising results for full mesh pose recovery from videos.

\section*{Acknowledgment}
This research was sponsored by the Australian Research Council (ARC) grant DP160101458 and partially supported by ARC grant DP190102443. The Tesla K-40 GPU used for this research was donated by the NVIDIA Corporation.

%\ifCLASSOPTIONcaptionsoff
%  \newpage
%\fi



\bibliographystyle{IEEEtran}
\bibliography{IEEEabrv,ijcai19}

\balance

\end{document}